\documentclass[pdflatex,sn-mathphys-num]{sn-jnl}

\usepackage{graphicx}%
\usepackage{multirow}%
\usepackage{amsmath,amssymb,amsfonts}%
\usepackage{amsthm}%
\usepackage{mathrsfs}%
\usepackage[title]{appendix}%
\usepackage{xcolor}%
\usepackage{textcomp}%
\usepackage{manyfoot}%
\usepackage{booktabs}%
\usepackage{algorithm}%
\usepackage{algorithmicx}%
\usepackage{algpseudocode}%
\usepackage{listings}%
\usepackage{anyfontsize}
\newenvironment{where}{\noindent{}where\begin{itemize}}{\end{itemize}}

\begin{document}

\title[Article Title]{Reconstruction of dynamic systems using genetic algorithms with dynamic search limits}


\author[1]{\fnm{Omar } \sur{Rodríguez-Abreo}}\email{omar.rodriguez@upq.edu.mx}

\author[1]{\fnm{ José L. } \sur{Aragón}}\email{jlaragon@unam.mx}

\author*[1]{\fnm{ Mario A.} \sur{Quiroz-Juárez}}\email{maqj@fata.unam.mx}

\affil[1]{\orgdiv{Centro de Física Aplicada y Tecnología Avanzada}, \orgname{Universidad Nacional Autónoma de México}, \orgaddress{\street{Boulevard Juriquilla 3001}, \city{Querétaro}, \postcode{76230}, \state{State}, \country{México}}}


\abstract{Mathematical modeling is a powerful tool for describing, predicting, and understanding complex phenomena exhibited by real-world systems. However, identifying the equations that govern a system’s dynamics from experimental data remains a significant challenge without a definitive solution. In this study, evolutionary computing techniques are presented to estimate the governing equations of a dynamical system using time-series data. The main approach is to propose polynomial equations with unknown coefficients, and subsequently perform a parametric estimation using genetic algorithms. Some of the main contributions of the present study are an adequate modification of the genetic algorithm to remove terms with minimal contributions, and a mechanism to escape local optima during the search. To evaluate the proposed method, we applied it to three dynamical systems: a linear model, a nonlinear model, and the Lorenz system.  Our results demonstrate a reconstruction with an Integral Square Error below 0.22 and a coefficient of determination $R^2$ of 0.99 for all systems, indicating successful reconstruction of the governing dynamic equations.}




\maketitle

\section{Introduction}\label{intro}

Identifying the governing equations of natural phenomena is a significant challenge. Mathematical models may consist of partial differential equations or systems of first-order ordinary differential equations (ODEs) \cite{sundnes2002multigrid,jimenez2021experimental,quiroz2021reconfigurable}. In either case, a deep understanding of the physical laws governing the phenomenon is required to propose a model with a minimal set of differential equations \cite{debnath2005nonlinear,quiroz2019periodically,escobar2022classical}. Moreover, modeling often demands knowledge across multiple disciplines because the studied phenomena are usually multidisciplinary and the necessary information is not always available \cite{mate2}. Additionally, challenges such as time delays, noise inherent in system measurements, external disturbances, chaotic dynamics, algebraic loops, and parametric uncertainties further complicate this problem \cite{mate3}. In many cases, this traditional approach fails.

With the rapid advancements in data science and intelligent algorithms, there is a growing temptation to identify system models directly from data. This approach, known as data-driven modeling, has already been applied in various fields \cite{montans2019data,escobar2024data}, ranging from finance \cite{zhi2022managing} to healthcare \cite{rodriguez2024modeling,model3}. In \cite{Sindy}, a method called Sparse Identification of Nonlinear Dynamics (SINDy) was proposed to reconstruct parsimonious symbolic models directly from time-series data by identifying a sparse subset of terms from a candidate function library. In \cite{id2}, a method was proposed for automatically deriving symbolic differential equations from time-series data. This technique was successfully demonstrated across six different nonlinear coupled systems from various physical domains. Another example is presented in \cite{model1}, where an unmanned airship is modeled based on aerodynamic principles and mathematical analysis of the potential velocity flow around a vehicle’s shape. The entire system is described by nonlinear equations subject to boundary conditions governing air-structure interactions, and the model is obtained in an analytical environment. Additionally, \cite{model2} presented a new model, validated with experimental results, for the drug release process from polymeric microparticles using a dispersive fractal approximation of motion. Some studies can even describe phenomena without equations, such as the research in \cite{id1}, which used empirical dynamic modeling (EDM) for fisheries forecasts.  

As mentioned above, artificial intelligence algorithms have been used for modeling from massive datasets, primarily because they enable the analysis of complex data and the extraction of patterns that are difficult to identify through other methods \cite{AIR}. It is also worth mentioning the efforts to use deep learning techniques to identify dynamic systems from time-series data \cite{Deep,Deep2}, where artificial neural networks are employed. Despite the computing power available today, automating the process of finding differential equations that model nonlinear dynamic systems remains a challenge. The significant disadvantages of neural networks include their black-box nature, which means that no information can be inferred from the model only from its inputs and outputs, and the difficulty of incorporating known physical constraints. These limitations reduce the ability of AI-based systems to extrapolate dynamic behavior beyond the data with which the system is trained. Nonetheless, efforts have been made in other areas of AI to determine the governing equations of dynamic systems.

Some approaches do not rely on neural networks. For instance, genetic programming has been proposed as a method to obtain dynamic models \cite{rg}, where an algorithm uses symbolic regression to identify nonlinear dynamics. This research enables the derivation of models for a wide range of systems, from simple dynamic systems to more complex ones, such as chaotic double pendulums. However, symbolic regression has high computational costs, scales poorly for complex systems, and tends to overfit the models. Finally, some researchers have proposed using metaheuristic algorithms to derive dynamic models \cite{ianld}. These algorithms are commonly used for parameter estimation in known models \cite{Meta}, hyperparameter tuning in neural networks \cite{Meta2}, and as tools for identifying specific windows in time-series data \cite{Meta3}. Among metaheuristic algorithms, genetic algorithms are powerful tools that provide intelligent search capabilities, whereas heuristic algorithms have limitations \cite{Meta4}. Their main advantages include ease of implementation across various types of problems and the ability to efficiently handle noisy or incomplete data. Notably, genetic algorithms are among the most widely used metaheuristic techniques \cite{GAR}. However, they also have disadvantages such as the large number of hyperparameters required for optimal performance \cite{kramer2017genetic}. Despite this, they remain one of the most accepted metaheuristic algorithms by the scientific community.

In this study, we propose a method for identifying the governing differential equations by assuming a polynomial form. The polynomial is constructed from possible combinations of system variables, and the coefficients are determined through evolutionary computing, specifically using Genetic Algorithms (GA). Without prior knowledge of the systems to be modeled, the GA can struggle to solve the problem and is heavily dependent on the selection of hyperparameters, making the system particularly sensitive to the established search limits. To mitigate this, dynamic limits are introduced in this study to avoid such issues. Additionally, two strategies were implemented to escape local optima and eliminate terms with minimal contributions to the model.

To test the proposed method, we applied it to three dynamical systems: a linear model, a second-order nonlinear model, and the Lorenz system. The results show that all systems can be accurately modeled using polynomials fitted with genetic algorithms. Specifically, the nonlinear dynamic system yielded a Root Mean Square Error (RMSE) of 0.00025 and $R^2$ of 0.99. The tests with the Lorenz attractor showed an RMSE of 0.33 and an $R^2$  of 0.99. The implementation of simple GAs without dynamic limit search is unable to reconstruct the dynamical equations from the Lorenz attractor when using the same hyperparameters and number of iterations as those employed in the dynamic limit method. This indicates that the dynamical equations of a given system can be reconstructed with high precision using the proposed method, resulting in a parsimonious dynamic model.

\section{Materials and methods}

A dynamical system is defined by a set of ordinary differential equations,
\begin{equation}
  \label{ds}
 \frac{d}{dt} {\mathbf x} (t) = {\mathbf f} ( {\mathbf x}(t) ),
\end{equation}
where $\mathbf{x}(t) = \left( x_1 (t), x_2 (t), \ldots, x_n (t) \right)^T$ is the transpose of a vector that describes the state of the system at time $t$, and ${\mathbf f} ( {\mathbf x}(t) ) = \left( f_1 ( {\mathbf x}(t) ), f_2 ( {\mathbf x}(t), \ldots, f_n ( {\mathbf x}(t) ) )\right)^T$ contains the functions that describes the dynamics of the system. Formally, a dynamical system is a map obtained from the solution of \eqref{ds},  with an adequate initial condition \cite{mate1}.  However, for simplicity, we call the dynamical system a set of linear or nonlinear first-order differential equations, such as \eqref{ds}. As discussed above, to determine the function ${\mathbf f} ( {\mathbf x}(t) )$, knowledge of the physical laws involved and their mathematical formulation is required. In many cases, this is not possible. Thus, our purpose is to determine ${\mathbf f} ( {\mathbf x}(t) )$ from the experimental data. 

To address this, we assume that a time series for each variable of the system response is available, and we propose that ${\mathbf f} ( {\mathbf x}(t) )$ has a polynomial structure with unknown coefficients, which are determined using genetic algorithms. Therefore the following ansatz is proposed 

\begin{equation}
  \label{func}
{\mathbf f} ( {\mathbf x} ) = \left[ 
\begin{array}{ll}
     P_1^m ( {\mathbf x} ) \\
     P_2^m ( {\mathbf x} ) \\
     \vdots \\
     P_n^m ( {\mathbf x} ) 
\end{array}
\right],
\end{equation}
where $P_i^m ( {\mathbf x} )$, $i=1,2, \ldots,n$, are polynomials in $\left( x_1 (t), x_2 (t), \ldots, x_n (t) \right)$ up to $m$-th order. If the coefficients of each $P_i^m$ are $\boldsymbol{\xi}_i = \left( \xi_{i1}, \xi_{i2}, \ldots \right)^T$ then the matrix
\begin{equation}
 \label{pol}
\boldsymbol{\Xi} = \left[ \boldsymbol{\xi}_1 \boldsymbol{\xi}_2 \cdots \right],
\end{equation}  
contains the unknown coefficients that determine which terms of each $P_i^m$ are active in the dynamics. Using genetic algorithms, we aim to find the sparse matrix $\boldsymbol{\Xi}$ from time series.

For comparison purposes, we will use the same three examples that were solved in \cite{Sindy}, including a second-order linear model, a second-order nonlinear model, and the Lorenz system. The first one is the second-order linear system given by: 
\begin{equation}
 \label{oscLin}
\begin{array}{rcl}
  \dot{x}&=&-0.1x+2y ,\\
 \dot{y}&=&-2x-0.1y ,\\
\end{array}
\end{equation}
and the second is a cubic system:
\begin{equation}
 \label{oscCub}
\begin{array}{rcl}
  \dot{x}&=&-0.1x^3+2y^3 ,\\
 \dot{y}&=&-2x^3-0.1y^3 .\\
\end{array}
\end{equation}
Now, suppose that the dynamical systems \eqref{oscLin} and \eqref{oscCub} are unknown, but the corresponding time series $x$ and $y$ are available. The goal is to reconstruct the dynamical equations of the system using these signals.

As described above, we assume that \eqref{oscLin} and \eqref{oscCub} can be expressed as linear combinations of the polynomial functions in \eqref{func}. For both examples, we propose third-order polynomials:
\begin{equation}\label{polO}
\begin{array}{rcl}
  \dot{x}&=&a_1x+a_2x^2+a_3x^3+a_4y+a_5y^2+a_6y^3+a_7xy^2+a_8yx^2+a_9xy ,\\
 \dot{y}&=&b_1x+b_2x^2+b_3x^3+b_4y+b_5y^2+b_6y^3+b_7xy^2+b_8yx^2+b_9xy .\\
\end{array}
\end{equation}
The matrix of unknown coefficients
\begin{equation}
 \label{eqn:coefs12}
 \boldsymbol{\Xi} = \left[ 
    \begin{array}{ll}
         a_1 & b_1  \\
         a_2 & b_2  \\
         a_3 & b_3  \\
         a_4 & b_4  \\
         a_5 & b_5  \\
         a_6 & b_6  \\
         a_7 & b_7  \\
         a_8 & b_8  \\
         a_9 & b_9  
    \end{array}
    \right] ,
\end{equation}
will be determined from the data for each case \eqref{oscLin} and \eqref{oscCub}.

The last example is a particular case of the well-known Lorentz system:
\begin{equation}
  \label{Lorenz}
\begin{array}{rcl}
  \dot{x}&=&10(y-x) ,\\
 \dot{y}&=&x(28-z)-y ,\\
 \dot{z}&=&xy-\frac{8}{3} z .
\end{array}
\end{equation}

For this case, we also propose a third-order polynomial expansion, which can be written as:
\begin{equation}
 \label{polL}
\begin{array}{rcl}
\dot{x}&=&a_1x+ a_2x^2 + a_3x^3+ a_4y+a_5y^2+a_6y^3+a_7z+a_8z^2+a_9z^3+a_{10}x^2y+a_{11}x^2z+ \\
&&+a{12}y^2x+a_{13}y^2z+a_{14}z^2x+a_{15}z^2y+a_{16}xyz+a_{17}xy+a{18}xz+a_{19}xz,\\
\dot{y}&=&b_1x+ b_2x^2 + b_3x^3+ b_4y+b_5y^2+b_6y^3+b_7z+b_8z^2+b_9z^3+b_{10}x^2y+b_{11}x^2z+ \\
&&+b{12}y^2x+b_{13}y^2z+b_{14}z^2x+b_{15}z^2y+b_{16}xyz+b_{17}xy+b{18}xz+b_{19}xz,\\
\dot{z}&=&c_1x+ c_2x^2 + c_3x^3+ c_4y+c_5y^2+c_6y^3+c_7z+c_8z^2+c_9z^3+c_{10}x^2y+c_{11}x^2z+ \\
&&+c{12}y^2x+c_{13}y^2z+c_{14}z^2x+c_{15}z^2y+c_{16}xyz+c_{17}xy+c{18}xz+c_{19}xz.
\end{array}
\end{equation}

The matrix of the unknown coefficient can be written as:
\begin{equation}
 \label{eqn:coefslorentz}
 \boldsymbol{\Xi} = \left[ 
    \begin{array}{lll}
         a_1 & b_1 & c_1 \\
         a_2 & b_2 & c_2 \\
         a_3 & b_3 & c_3 \\
         a_4 & b_4 & c_4 \\
         a_5 & b_5 & c_5 \\
         a_6 & b_6 & c_6 \\
         a_7 & b_7 & c_7 \\
         a_8 & b_8 & c_8 \\
         a_9 & b_9 & c_9 \\
         a_{10} & b_{10} & c_{10} \\
         a_{11} & b_{11} & c_{11} \\
         a_{12} & b_{12} & c_{12} \\
         a_{13} & b_{13} & c_{13} \\
         a_{14} & b_{14} & c_{14} \\
         a_{15} & b_{15} & c_{15} \\
         a_{16} & b_{16} & c_{16} \\
         a_{17} & b_{17} & c_{17} \\
         a_{18} & b_{18} & c_{18} \\
         a_{19} & b_{19} & c_{19} \\
    \end{array}
    \right].
\end{equation}

\subsection{Genetic algorithms as parametric estimators}
 
Genetic algorithms are used to estimate the parameters \eqref{eqn:coefs12} of the ansatz \eqref{polO} for systems \eqref{oscLin} and \eqref{oscCub}, and in the case of the coefficients \eqref{eqn:coefslorentz} of the ansatz \eqref{polL} for the Lorentz system \eqref{Lorenz}.

Algorithm \ref{JO} shows the pseudocode for the GA used in this study.

\begin{algorithm}
	\caption{A Genetic Algorithm Pseudocode \label{JO}}
	\begin{algorithmic}[1]
        \State \textbf{Begin}
        \State \textbf{Set hyperparameters}
	 \State Create an initial random population;
		\While {Iterations$_{count}$ $ \leq$ Gen (Gen=Generations number)}
		\State Calculate the fitness function for each individual.
		\State Select the best individual (with lower fitness value), according to Wheel Roulette Selection and biological pressure.
		\State Crossing the best individuals; the division is created using a random single point; two children are created.
		\State Mutation operator is applied.
		\State New populations are created with children, maintaining the best solutions (elitism)
		\EndWhile
		\State Save the individual with the lowest fitness value
		\State \textbf{end}
	\end{algorithmic} 
\end{algorithm}

 GA starts with random solutions that are evaluated using a cost or fitness function, which is usually a statistical parameter. The most common fitness functions include mean squared error (MSE), root mean squared error (RMSE), coefficient of determination ($R^2$), and integral square error (ISE). In this study, the ISE defined by \eqref{eq:ISE} is used as an algorithm performance indicator because it allows us to determine the goodness of all quadratic indicators (maximize significant errors and minimize minor errors) and enables penalizing errors throughout the entire signal. However, MSE \eqref{eq:MSE} and ($R^2$) \eqref{eq:R2} are calculated for the final dynamic models obtained.

\begin{eqnarray}
ISE &=& \int (v_i-\hat{v}_i)^2dt , \label{eq:ISE} \\
MSE &=& \sum_{i=1}^n (v_i-\hat{v}_i)^2 , \label{eq:MSE} \\
R^2 &=& 1-\frac{\sum_{i=1}^n (v_i-\hat{v}_i)^2}{\sum_{i=1}^n (v_i-\overline{v})^2}, \label{eq:R2}
\end{eqnarray}
\begin{where}
  \item $v_i$ = actual value in the time series.
  \item $\hat{v_i}$ = estimated value by the algorithm.
  \item $\overline{v}$ = the mean value of the variable.
  \item $n$ = number of samples in the time series.
\end{where}

 To obtain the real data used to estimate the dynamical system, numerical integrations of  \eqref{oscLin}, \eqref{oscCub}, and \eqref{Lorenz} were performed with initial conditions $[x,y]^T=[0 ,2]^T$ for \eqref{oscLin} and \eqref{oscCub}, and  $[x ,y, z]^T=[-8, 7 ,27]^T$ for the Lorentz system \eqref{Lorenz}. ODE45 numerical integration was used, and the results were considered as real, measurable data in a physical system. Numerical differentiation is necessary to obtain the derivatives of these variables. Using the $v_i$ values, the GA shown in pseudocode \ref{JO} can be executed. The algorithm uses elitism, Wheel Roulette Selection (WRS), mutation, and single-point crossover with double offspring. The selection pressure was set to 70\%, which reduced the reproduction probability of low-performance individuals. All the hyperparameter values used in our simulations are listed in Table \ref{TInputs}. 

\begin{table}[!ht]
\caption{\label{TInputs}Hyperparameters employed in the Genetic Algorithm. The initial population consists of vectors with random coefficients, biological pressure refers to the percentage of individuals that reproduce, and elitism is the percentage of individuals with the best performance that is preserved in each generation.}  

\begin{tabular}{|l|l|}
\hline
\textbf{Hyperparameter} & \textbf{Value} \\
\hline
Initial Population & 10,000 \\
Maximum search limit for \eqref{oscLin} and \eqref{oscCub} &  10 \\
Maximum search limit for \eqref{Lorenz} &  30  \\
Minimum search limit for all &  Negative of the upper limit \\
Number of generations & 2,000  \\
Fitness function &   $ISE$ \\
Biological pressure & 70\% \\
Mutation probability & 10\% \\
Elitism & 10\% \\
\hline
\end{tabular}
\end{table}

 The GA with the hyperparameters shown in Table \ref{TInputs} was used to identify the coefficients of models \eqref{eqn:coefs12} and \eqref{eqn:coefslorentz}. Although the proposed models \eqref{Lorenz} and \eqref{polL} have low-magnitude error values, they contain eighteen and fifty-seven terms, respectively. In contrast, real dynamic models usually have only a few terms that represent most system dynamics. Therefore, it is necessary to integrate a mechanism to eliminate the terms with minimal relevance to the model. 

\subsection{Dynamic limits in Genetic Algorithms}

One of the major limitations of genetic algorithms is the correct selection of search hyperparameters. Some hyperparameters, such as the mutation rate or selection pressure, can directly affect the search speed or the algorithm's ability to escape local optima. However, the choice of search limits is a critical issue; without prior information, it is impossible to define the upper and lower search limits. In known problems, the expected magnitude values or the physical conditions of the phenomenon are often used. However, if the search limits are not correctly chosen, the algorithm may fail to find adequate solutions. Because the algorithm cannot find values outside the fixed search range, it may compensate by increasing the magnitudes of other low-relevance terms, giving the appearance that all terms are relevant for reconstructing the system dynamics. This often results in models with a large number of terms.  

To avoid models with numerous terms, instead of an equivalent model with a reduced number of terms, we propose a strategy based on dynamic limits, which represents an improvement over the standard GA. The proposed strategy essentially consists of: if the determined value of some coefficient is close to the maximum search limit or, in contrast, the limit is very far from the estimated value, then must be adjusted. If the value of a coefficient determined by the GA exceeds 90\% of the search limit's magnitude, the limits increase by 10\%. Conversely, if the coefficient value is less than 90\% of the limit, the limit is reduced by 10\%.

This procedure causes the limits to align more closely with the estimated coefficient values, allowing values with minimal contributions to trend toward zero more quickly. However, this also significantly increases the likelihood of becoming trapped in the local optima. Although some coefficients may have small magnitudes, it is unlikely that the genetic algorithm will assign a value of zero. Therefore, a minimum threshold was defined. If the coefficient value falls below this threshold, the coefficient is set to zero and excluded from the search. Once the coefficient is determined as zero, the original search limits are restored. This allows the search to restart in a broad space, but with fewer coefficients, enabling the algorithm to escape local optima.

\begin{figure}[!ht]
\centering
\includegraphics[width=0.75\linewidth]{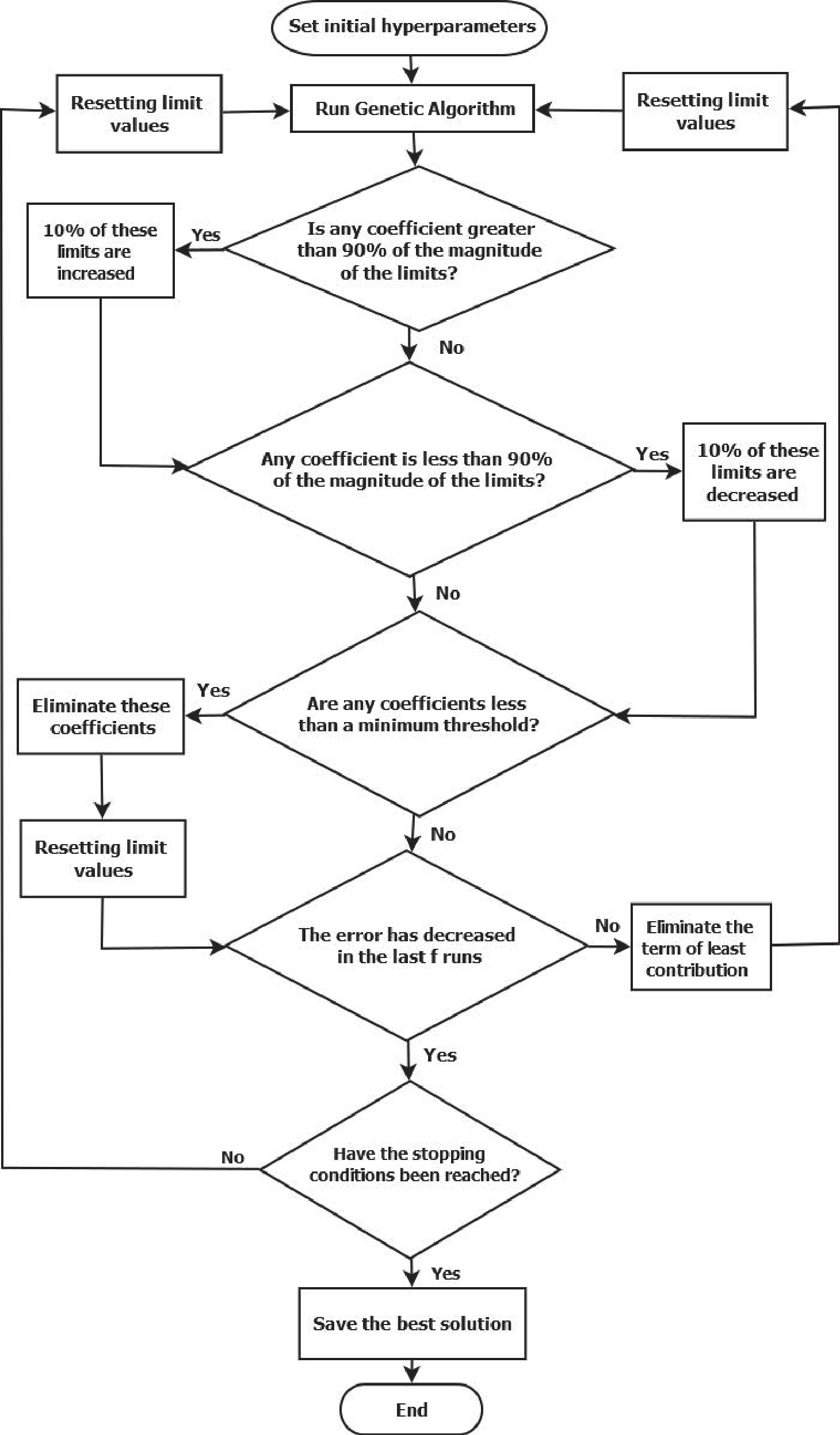}
\caption{ General scheme of the proposed strategy for implementing dynamic limits for the genetic algorithms.}
\label{EGLD}
\end{figure}

Another strategy to avoid local optima is to eliminate the coefficient that contributes less to the reconstruction of the signal in the sense that no reduction in the minimum value of the fitness function is observed after running a certain number 
of iterations, $f$. The algorithm continues to reevaluate the error without this coefficient. If the error decreases after $f$ iterations, the change is considered valid; otherwise, the next smallest coefficient is set to zero and the process is repeated. Each time a coefficient is removed, the search limits are reset to their original value. A general scheme for implementing the dynamic search limits and automatic coefficient removal is shown in Figure \ref{EGLD}.

 A series of GAs were run with hyperparameters similar to those listed in Table \ref{TInputs}. For a fair comparison, the number of fitness function evaluations was maintained constant. However, under the approach proposed in this study, some adjustments in hyperparameter selection are required. It is essential to select a threshold such that, if any coefficient is below this threshold, the GA treats the coefficient as zero. Additionally, it is necessary to define the stagnation criterion, that is, the number of iterations without a reduction in the fitness value, after which the GA is considered to have reached a local optimum. To overcome this, the term with the lowest magnitude is discarded. The final hyperparameter values for the proposed GA, including the threshold and stagnation values, are presented in Table \ref{Hyp2}.

\begin{table}[!ht]
\caption{\label{Hyp2}Hyperparameters employed to define dynamics limits in genetic algorithms. The stagnation refers to the Maximum number of genetic algorithms allowed without change in the value of the fitness function.}
\begin{tabular}{|p{3.5cm}|p{6.5cm}|}
\hline
\textbf{Hyperparameters} & \textbf{Value}\\
\hline
Iterations & 2000 \\
Initial population & 100 \\
Generations & 100 \\
Initial upper search limit & Maximum of the real signal divided by the polynomial term corresponding to the coefficient under consideration\\
Initial lower search limit &  The negative of the upper search limit \\
Threshold &   $1\times10^{-4}$ \\
Stagnation (\textit{f}) & 20 \\
\hline
\end{tabular}
\end{table}

 These hyperparameters allow a more efficient search for each GA compared to GAs that consider fixed limits. The hyperparameters for iterations, population size, and generations were chosen to ensure an equal number of fitness function evaluations for both the genetic algorithm with dynamic and fixed limits. This allowed for a comparison of both strategies at the same computational cost.

\section{Results}

In this Section, the results obtained for the reconstruction of the three considered dynamical systems are presented. For comparison, a GA with fixed and dynamic search limits was used. 

\subsection{GA with fixed search limits}

\begin{figure*}[!ht]
\begin{tabular}{c c}
\centering
  \includegraphics[width=.5\textwidth]{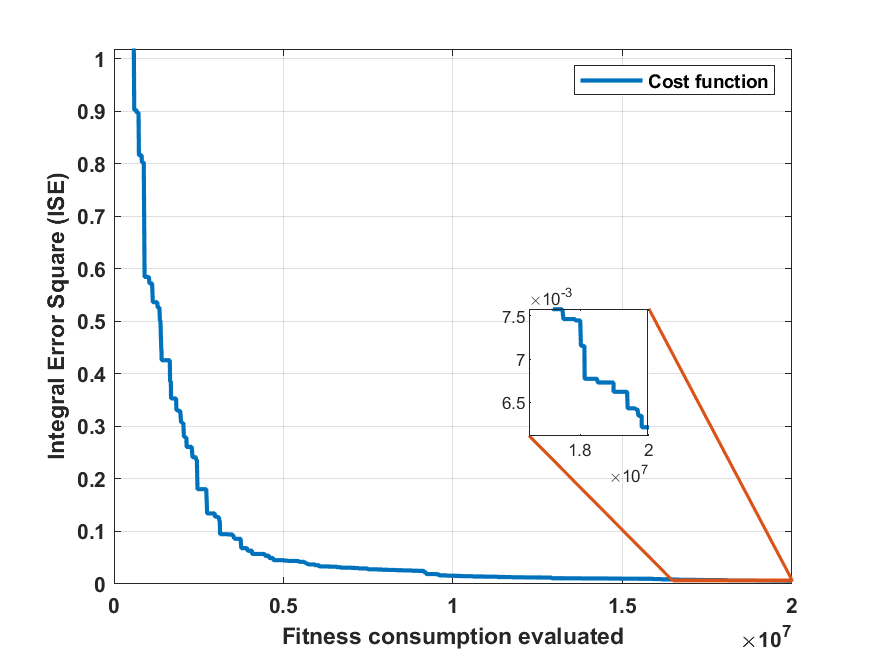}&
  \includegraphics[width=0.5\textwidth]{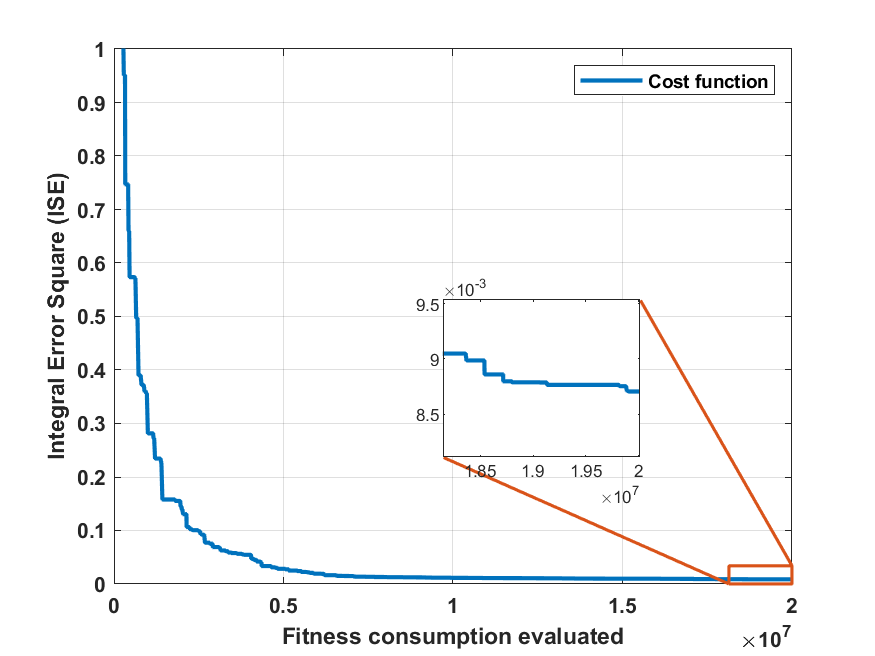}\\
  a)& b)
\end{tabular}
\caption{Performance of the GA for the reconstruction of the linear a) and nonlinear dynamical system\eqref{oscCub}.}
\label{PerOsc}       
\end{figure*}

For the linear \eqref{oscLin} and nonlinear \eqref{oscCub} dynamical systems, represented by third-order polynomials \eqref{polO}, and with the hyperparameters in Table \ref{TInputs} the GA gives the estimated coefficients listed in Table \ref{ResOscC}, and the performance of the algorithm is displayed in Figure \ref{PerOsc}.

\begin{table}[!ht]
\caption{\label{ResOscC}The coefficients obtained by the GA with the hyperparameters in Table \ref{TInputs} for the linear \eqref{oscLin} and nonlinear \eqref{oscCub} dynamical systems with fixed search limits.}
\begin{tabular}{c|c|ccccccccc}
Case  &n& 1 & 2 & 3 & 4 & 5 & 6 & 7 & 8 & 9\\
\hline\rule{0pt}{12pt}
linear  &$a_n$& 0.0627 & 0.0701 &-0.0809& 1.9605 & -0.0264 & 0.0191 & -0.0096 &-0.0566&0.00209\\
 &$b_n$ &-1.5920 &0.09109 & -0.1771 & -0.1624&-0.0637&0.035661&-0.0782&-0.2253&-0.0127\\
\hline
non-  &$a_n$& -0.1623 & -0.0385 &-0.0301& -0.1544 & 0.0385 & 2.0588 & 0.1035 &0.1199&0.0967\\
linear &$b_n$ & -0.5463& 0.5462&-0.05773&-1.8082&0.3324&-0.0604&-0.2347&0.0910&0.1498\\
 \hline
\end{tabular}
\end{table}

\begin{figure*}[!ht]
\begin{tabular}{c c}
\centering
  \includegraphics[width=.5\textwidth]{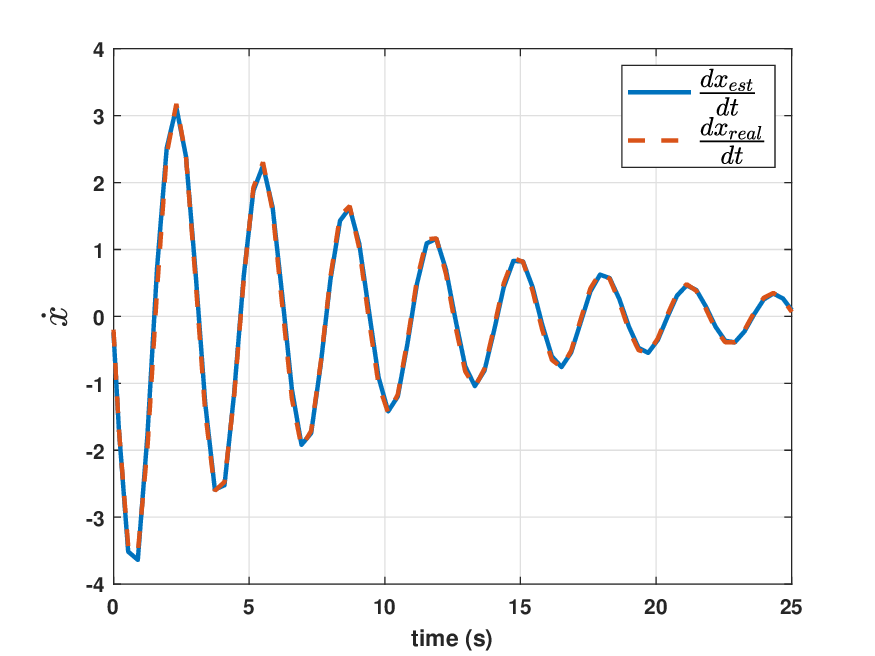}&
  \includegraphics[width=0.5\textwidth]{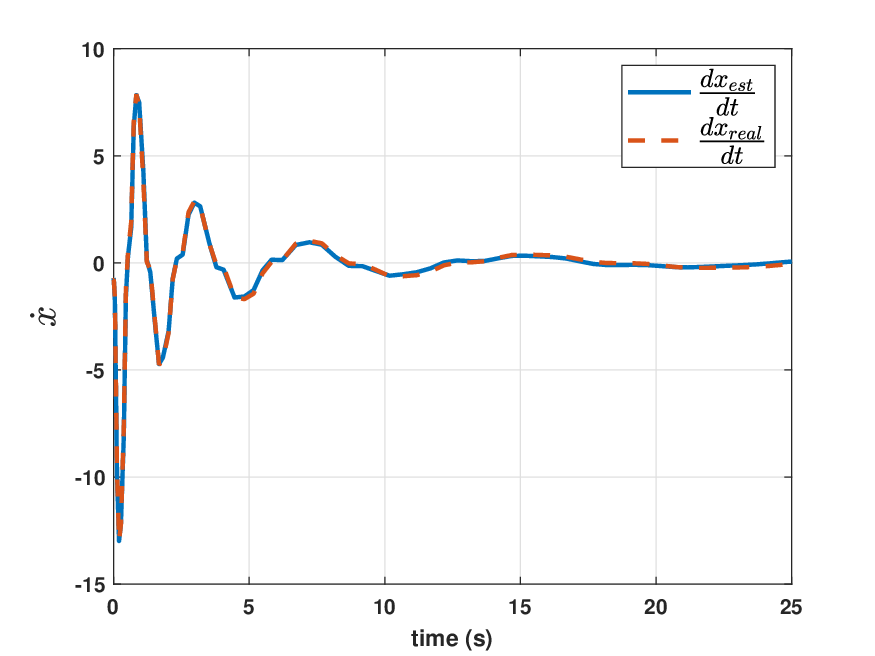}\\
    a)& b)\\
    \includegraphics[width=.5\textwidth]{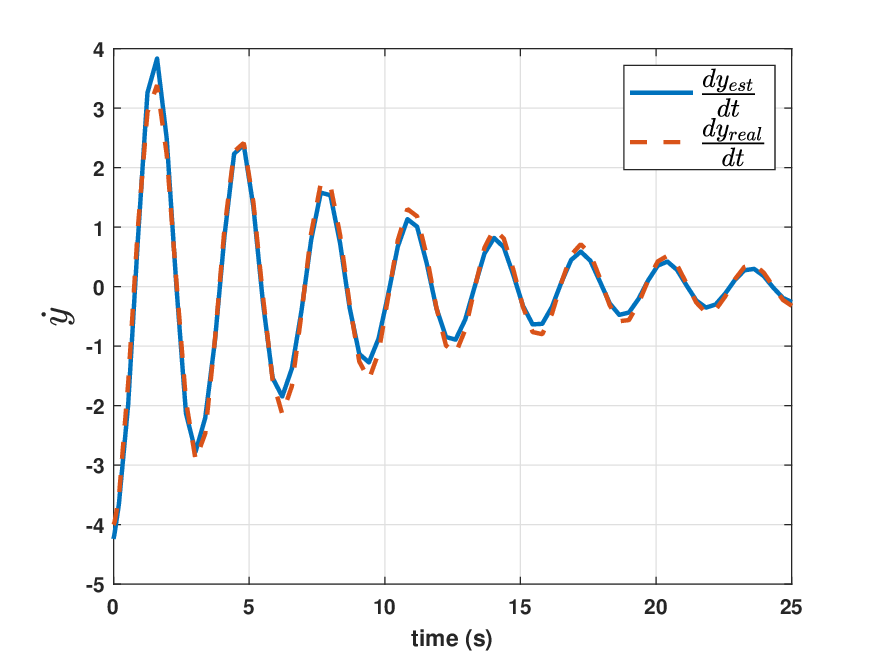}&
  \includegraphics[width=0.5\textwidth]{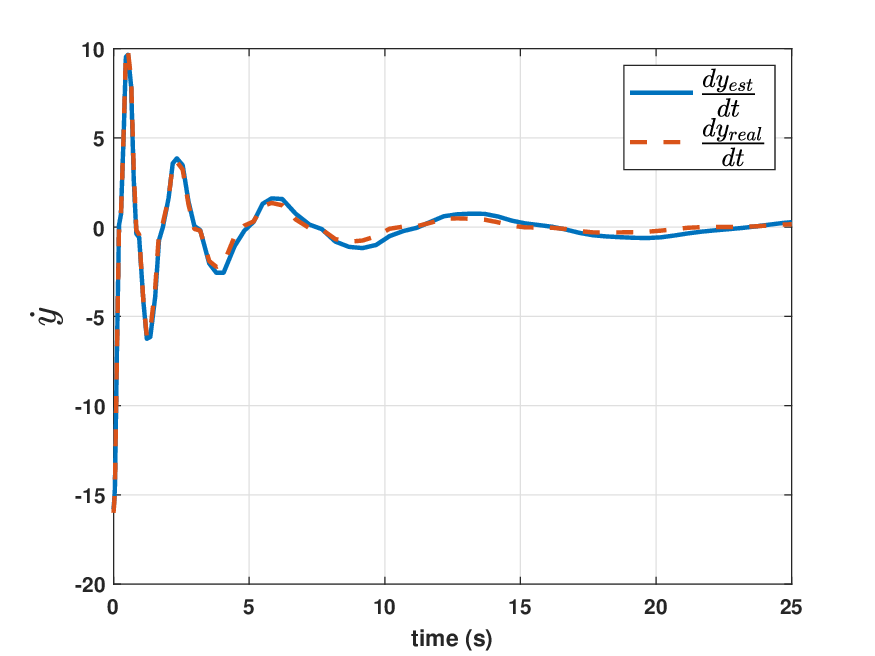}\\
  c)& d)
\end{tabular}
\caption{Comparison between the real values, $\dot{x}$ and $\dot{y}$, and those estimated by the GA, $\dot{x}_{est}$ and $\dot{y}_{est}$. a) and c) for the linear system \eqref{oscLin}, and b) and d) for the nonlinear system \eqref{oscCub}.} 
\centering
\label{OscDer}       
\end{figure*}

With the coefficients in Table \ref{OscDer}, for the linear case, $\dot{x}$ is reproduced with ISE$=0.00038$, MSE$=0.005$, and a coefficient of determination $R^2 = 0.9966$. By contrast, $\dot{y}$ is reproduced with ISE$=0.0019$, MSE$=0.0247$, and $R^2 = 0.9905$. For the nonlinear case,  $\dot{x}$ was reproduced with ISE$=0.0012$, MSE$=0.0154$, and $R^2 = 0.9984$. In contrast, $\dot{y}$ is reproduced with ISE$=0.005$, MSE$=0.0657$, and $R^2=0.9976$. The comparisons between real and estimated $\dot{x}$ and $\dot{y}$ are shown in Figure  \ref{OscDer}. 

\begin{figure*}[!ht]
\begin{tabular}{c c}
\centering
  \includegraphics[width=.5\textwidth]{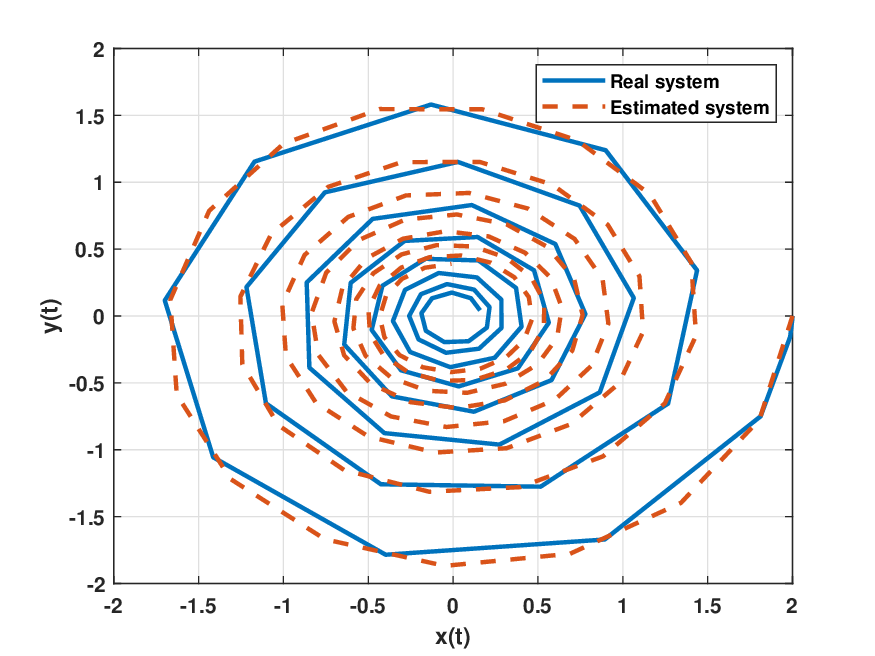}&
  \includegraphics[width=0.5\textwidth]{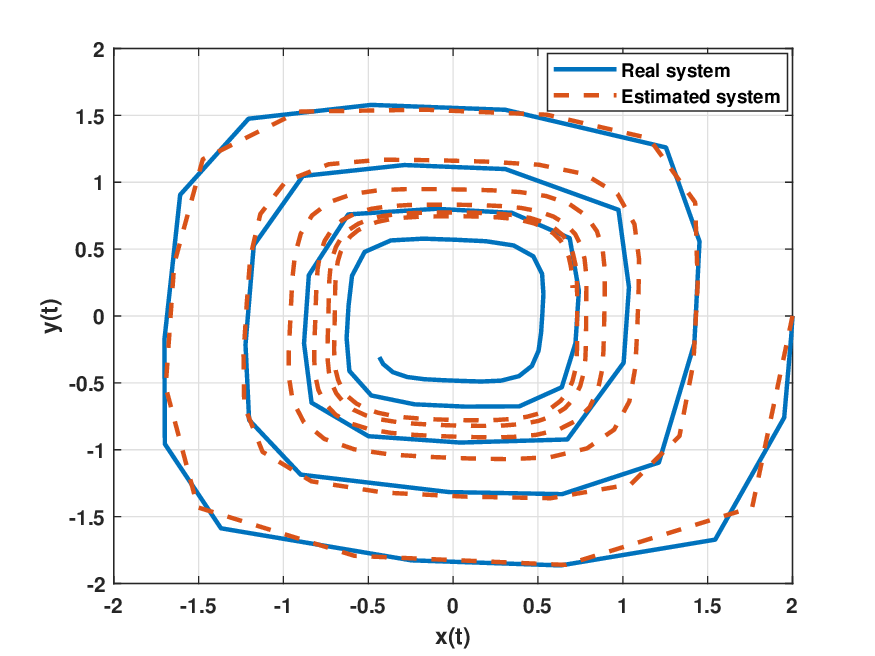}\\
  a)& b)
\end{tabular}
\caption{Comparisons of real and estimated signals for a) the linear system, and  b) the nonlinear system.}
\label{vs1}      
\end{figure*}

To obtain the signals $x$ and $y$, the estimated equations for $\dot{x}_{est}$ and $\dot{y}_{est}$ are solved using ODE45, and the results are shown in Figure \ref{vs1}.

\begin{figure}[!ht]
\centering
\includegraphics[width=0.7\linewidth]{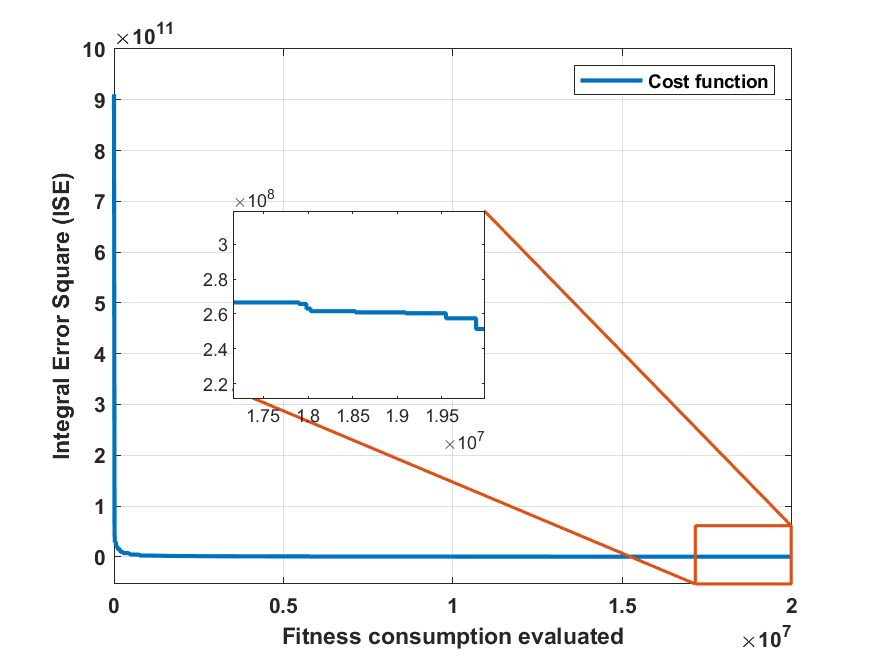}
\caption{Performance of the genetic algorithm in the reconstruction of the Lorenz system.}
\label{PerLor}
\end{figure}

Now, we apply the GA to reconstruct the Lorenz system \eqref{Lorenz} with the hyperparameters in Table \ref{TInputs} with fixed search limits. The obtained coefficients are shown in Table \ref{ResLor}, and the performance of the algorithm is shown in Figure \ref{PerLor}.

\begin{table}[!ht]
\caption{\label{ResLor}Coefficients obtained through the Genetic Algorithm with fixed limits for the Lorenz system}
\begin{tabular}{c|ccccccccccc}
n  & 1 & 2 & 3 & 4 & 5 & 6 & 7 & 8 & 9 &10\\
\hline\rule{0pt}{12pt}
$a_n$  & -12.5763 &	-20.8864 &	-26.6121 &	18.1858 &	-4.3300&	1.2810 &20.3331&	5.9778&	-0.9161 &	28.8342 \\
$b_n$  & -25.7636&	-19.8113&	13.7910&	-14.7880& 10.1368 &	-16.8971 &	-24.6484 &	-29.7303 &	1.3855  &	4.2580\\
$c_n$  & -20.7462 &	-21.4323 &	14.7666 &	18.9765 &	1.7754 &	-1.2818 &	-3.0809 &	23.9992 &	-1.7336 &	-4.0413\\
\hline
n  & 11 & 12 & 13 & 14 & 15 & 16 & 17 & 18 & 19 \\
\hline
$a_n$  & 9.1714 &	-11.9126 &	-0.4140 &	2.9400 &	-0.1346&	-3.7573 &0.1188 &-25.1999&-17.9396\\
$b_n$  & -4.3178&	29.0791&	0.8143&	-0.5718&-5.6551 &	0.3831 &	13.8218  &	14.2295 & 28.7571\\
$c_n$  & 15.6325 &	1.0139 &	-0.5197 &	-2.3021&	0.1731 &	-7.3695 &	-5.7851 &	24.34764& -12.0116\\
\hline
\end{tabular}
\end{table}

In this example, we obtain $ISE \approx 2\times 10^8$, and thus, the GA was unsuccessful. Upon examining the cost function, it was observed that, although the fitness function decreased, the problem could not be solved within the iterations used. In the following section, we apply the GA with dynamic limits, which can be used to solve these and general nonlinear dynamical systems. 

\subsection{Results for GA with dynamic limits}

In this section, we present the results of the GA with dynamic limits. For a better comparison, the times at which the fitness function is evaluated are the same as those in the genetic algorithm with fixed limits. The GA was applied to \eqref{oscLin} and \eqref{oscCub}, and the obtained coefficients are listed in Table \ref{ResOscLD}.

\begin{table}[!ht]
\caption{\label{ResOscLD}Coefficients obtained by the GA with dynamic limits for the linear \eqref{oscLin} and nonlinear \eqref{oscCub} dynamical systems.}
\begin{tabular}{c|c|ccccccccc}
Case  &n& 1 & 2 & 3 & 4 & 5 & 6 & 7 & 8 & 9\\
\hline\rule{0pt}{12pt}
linear  &$a_n$  & -0.099996 &	0 &	0 &	1.99999927 &  0&	0 &0 &	0 &	0 \\
 &$b_n$ &-2.000008 & 0 & 0 & -0.1000008& 0&0&0&0&0\\
\hline
non-  &$a_n$  & 0 &	0 &	-0.1000018 &	0 &  0&	1.999982 &0 &	0 &	0 \\
linear &$b_n$  & 0&	0&	-2.0000241 &	0 & 0 &	-0.099998 &	0 &	0 &	0\\
 \hline
\end{tabular}
\end{table}

\begin{figure*}[!ht]
\begin{tabular}{c c}
\centering
  \includegraphics[width=.5\textwidth]{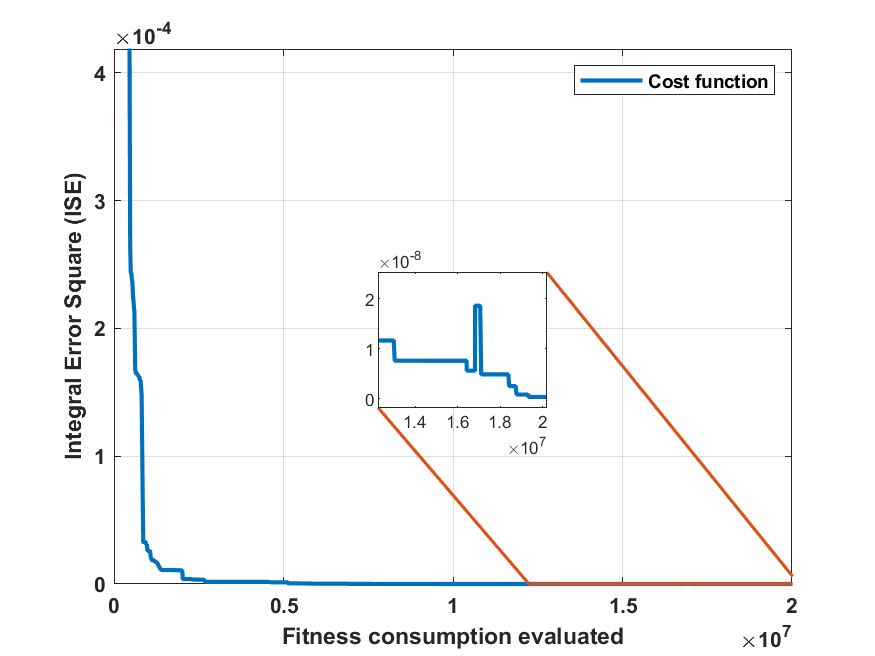}&
  \includegraphics[width=0.5\textwidth]{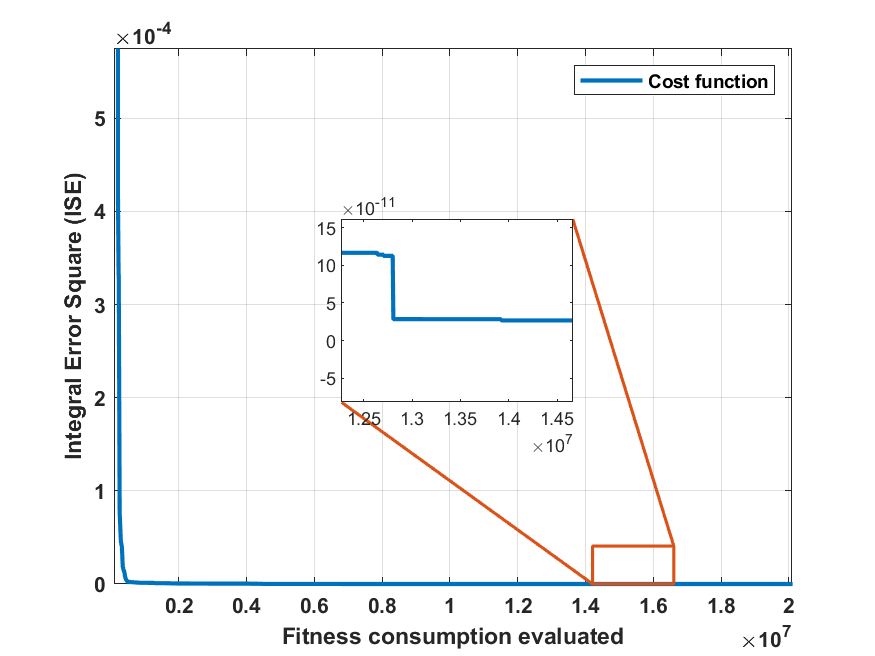}\\
  a)& b)
\end{tabular}

\caption{Performance of the GA with dynamic limits for the linear a) and nonlinear b) dynamical system.}
\label{PerOscAut}
\end{figure*}

The performance of the algorithm with dynamic limits is shown in Fig. \ref{PerOscAut}.

\begin{figure*}[!ht]
\begin{tabular}{c c}
\centering

  \includegraphics[width=0.5\textwidth]{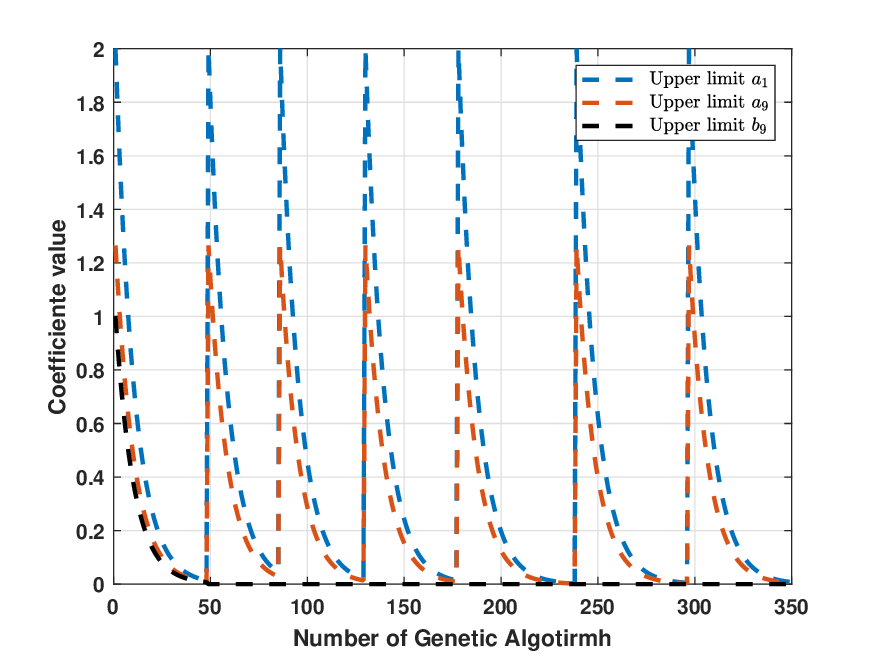}&  
 \includegraphics[width=0.5\textwidth]{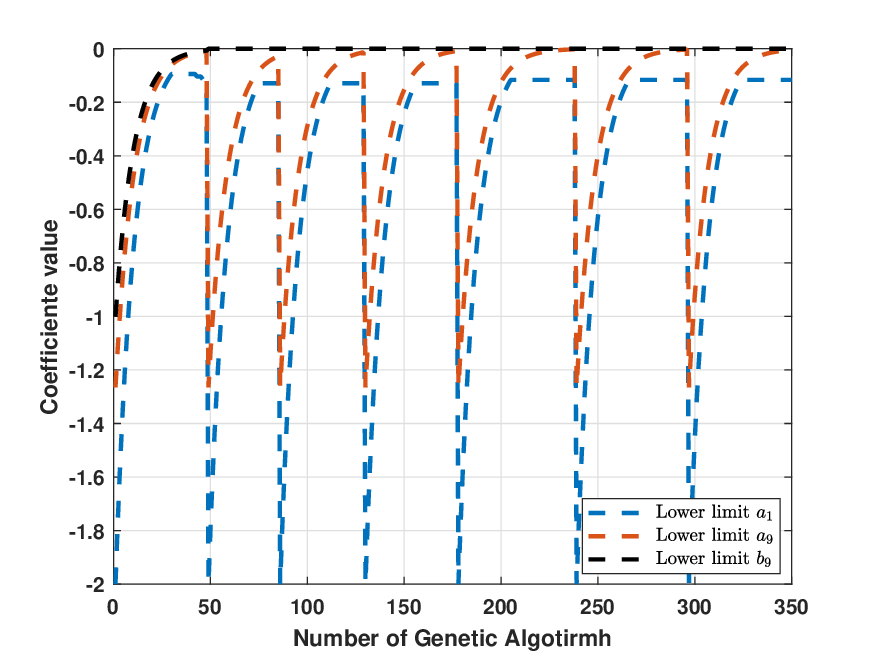}\\ a) &  b) \\

     \includegraphics[width=0.5\textwidth]{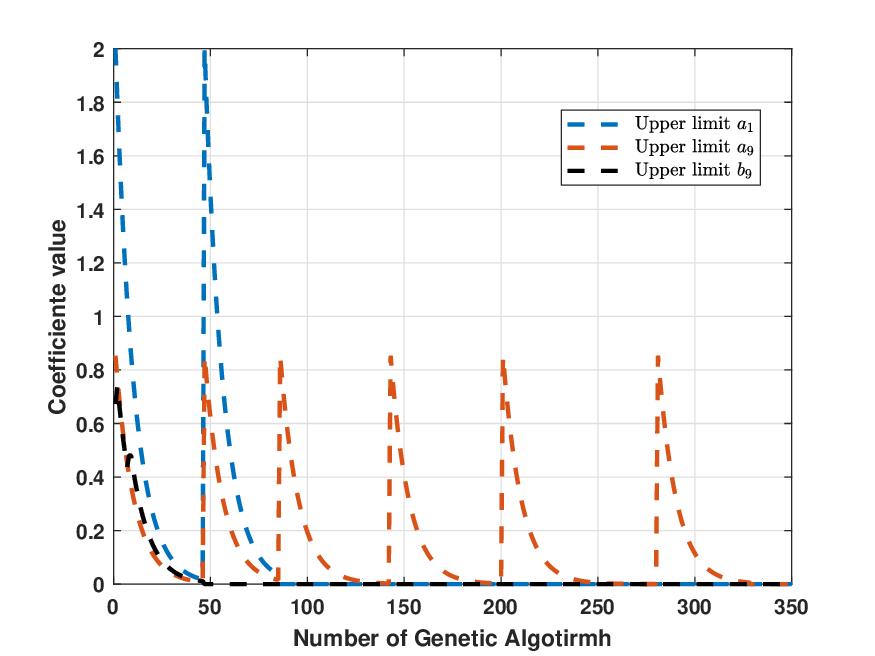}&   
\includegraphics[width=0.5\textwidth]{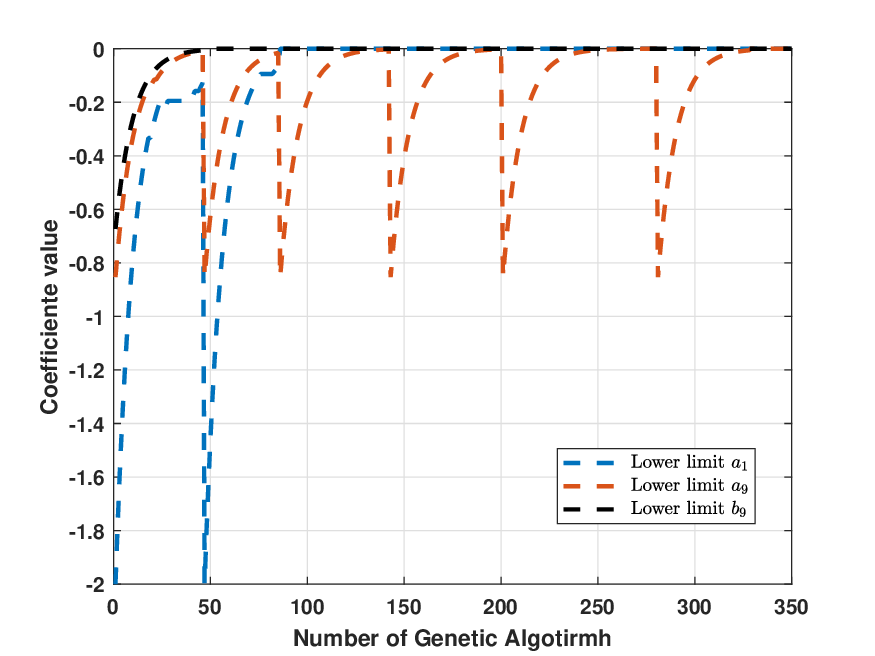}\\    c) & d)\\
\end{tabular}
\caption{Evolution of the dynamic upper and lower search limits for the linear system a) and b), and for the nonlinear system c) and d).}
\label{DL}      
\end{figure*}

It should be noted that the error sometimes increases owing to the strategies employed to break the local optima. Additionally, Fig. \ref{DL} a) and b) illustrate the evolution of the dynamic upper and lower search limits, respectively, for some coefficients in the linear case and Figures \ref{DL} c) and d) the evolution of upper and lower limits.

\begin{figure*}[!ht]
\begin{tabular}{c c }
\centering
  \includegraphics[width=.6\textwidth]{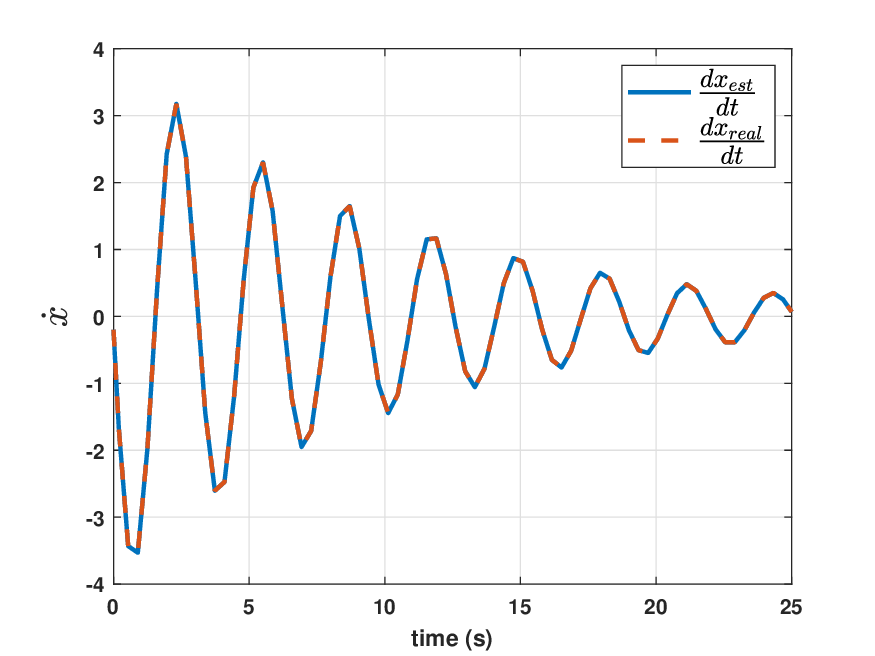}&
  \includegraphics[width=0.6\textwidth]{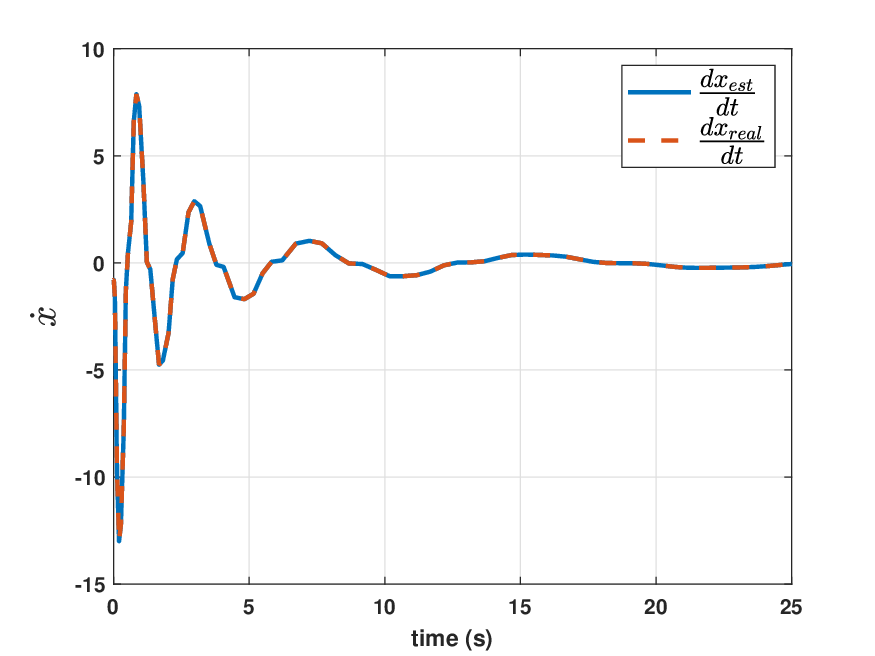}\\
  a)& b)\\
    \includegraphics[width=.6\textwidth]{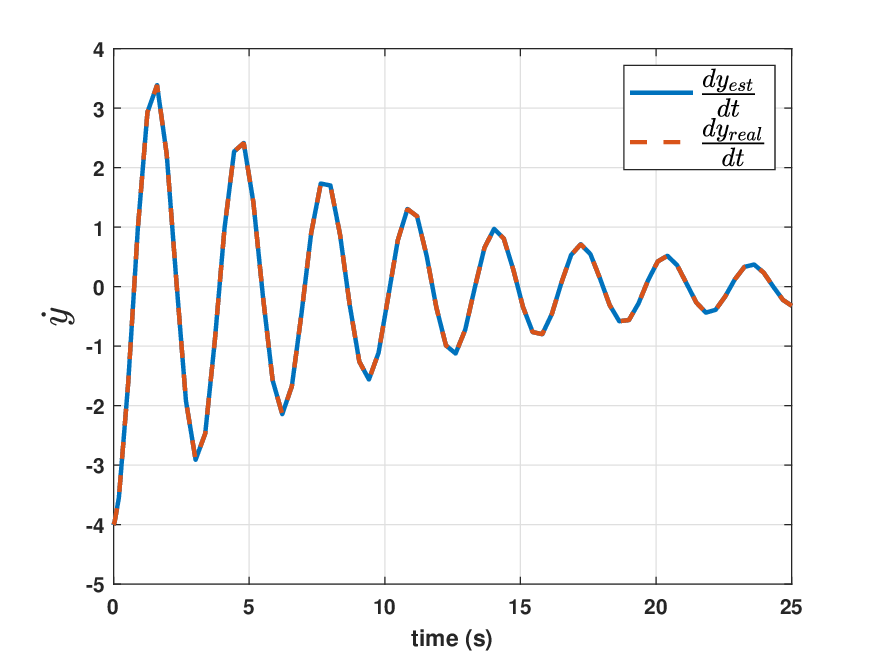}&
  \includegraphics[width=0.6\textwidth]{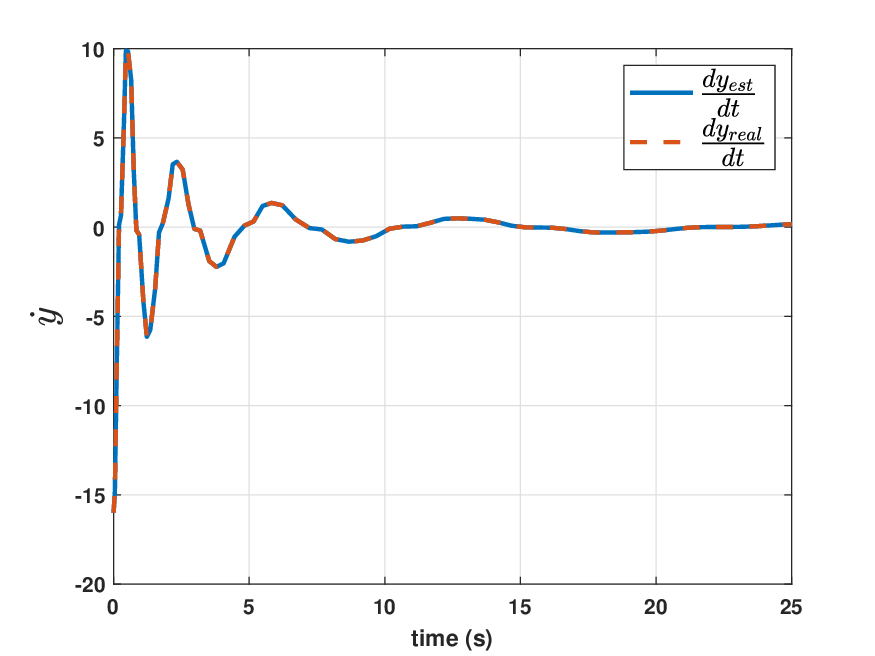}\\
  c)& d)
  
\end{tabular}

\caption{Comparison of real data and estimated using GA with dynamic limits. Real and estimated values of $\dot{x}$ for a) linear and b) nonlinear systems. Similarly, the comparison $\dot{y}$ is plotted for c) linear system and d) nonlinear system.}
\label{LDORvsOE}      
\end{figure*}

\clearpage
The comparison of the real signals and those estimated by the GA with dynamic search limits is shown in Fig. \ref{LDORvsOE}.

Using the coefficients shown in Table 5, both the linear and non-linear cases were reconstructed. For the first case, $\dot{x}$ was reproduced with ISE$=5.7 \times 10^{-13}$, MSE$=7.71 \times 10^{-12}$, and $R^2 = 0.9999$, whereas $\dot{y}$ ISE$=3.2\times 10^{-12}$, MSE$=4.39 \times 10^{-11}$, and $R^2= 0.9999$. For the non-linear case, the reconstruction of $\dot{x}$ yield ISE$=6.0 \times 10^{-11}$, MSE$=7.8 \times 10^{-10}$, and $R^2 = 0.9999$, whereas for $\dot{y}$ ISE$=3.13 \times 10^{-10}$, MSE$= 4.31 \times 10^{-9}$, and $R^2= 0.9999$.

\begin{figure*}[!ht]
\begin{tabular}{c c}
\centering
  \includegraphics[width=.62\textwidth]{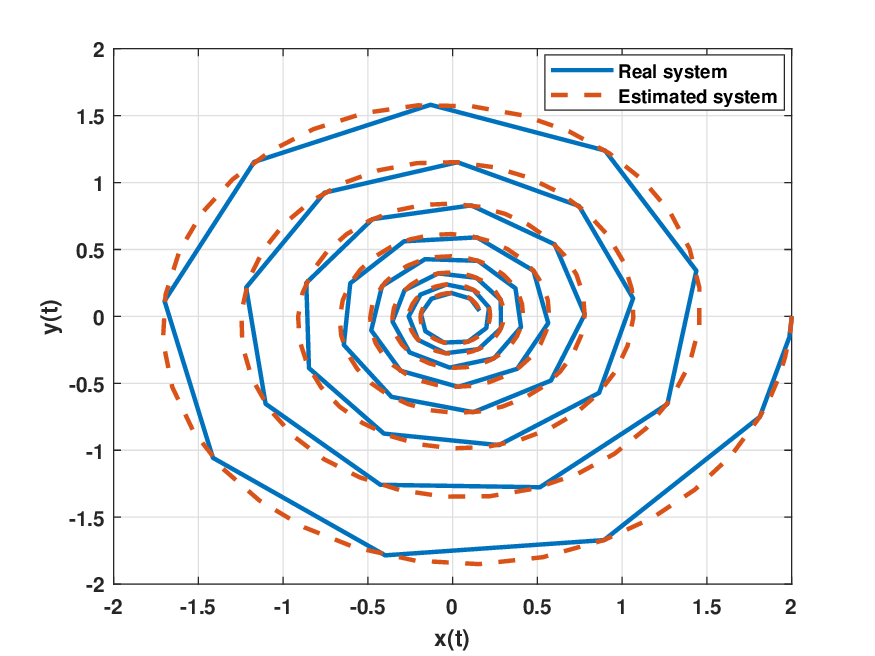}&
  \includegraphics[width=0.55\textwidth]{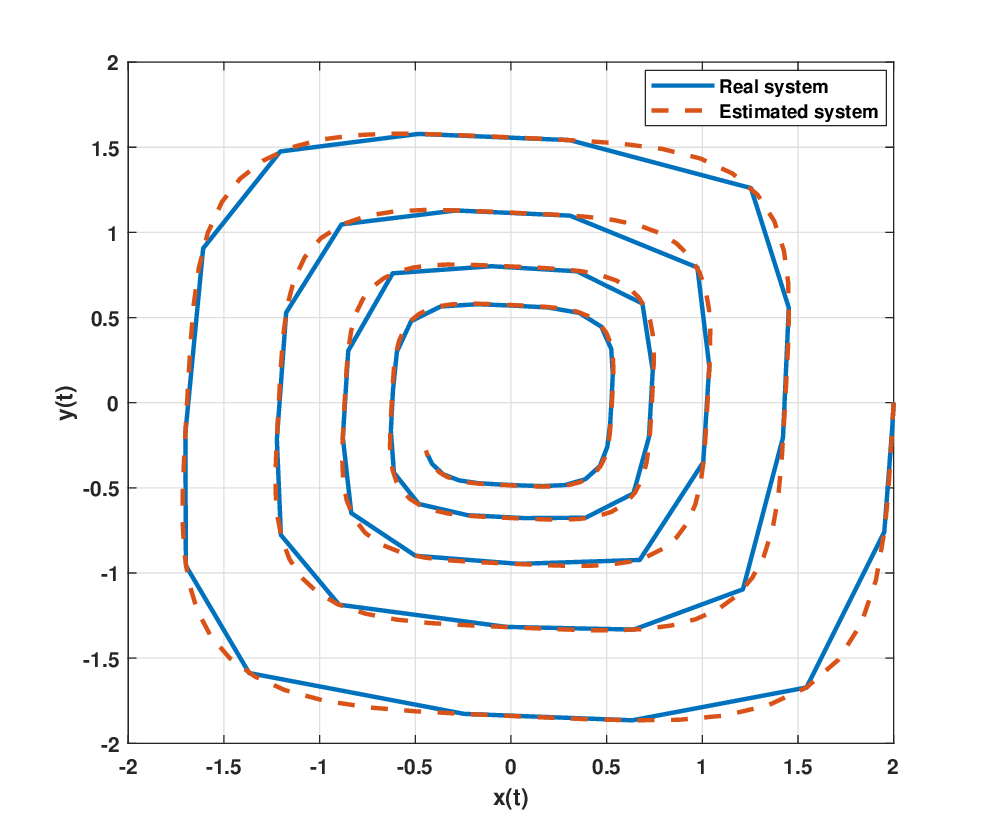}\\
  a)& b)
\end{tabular}
\caption{Comparison of the models obtained by the GA with dynamic search and the original systems. a) linear system and b) nonlinear system.}
\label{LDOAInt}
\end{figure*}

The next step is to perform a numerical integration of the estimated linear and nonlinear equations and compare them with the real signals in the phase space. ODE45 is used again and the results are shown in Figure \ref{LDOAInt}. 

 Finally, the proposed GA was implemented for Lorenz system reconstruction, and the obtained coefficients are listed in Table \ref{ResLorLD}.

\begin{table}[!ht]
\caption{\label{ResLorLD}Coefficients obtained by the GA with dynamics search limits for the Lorenz system}
\begin{tabular}{c|ccccccccccccccccccc}
n  & 1 & 2 & 3 & 4 & 5 & 6 & 7 & 8 & 9& 10 & 11& 12& 13& 14& 15& 16& 17& 18 &19\\
\hline\rule{0pt}{12pt}
$a_n$  & -9.9982&	0 &	9.9995 &	0 &	0 &	0 & 0 &	0 &	0 &	0 &0 &	0 &	0 &	0 &	0 &	0 &	0&	0 & 0\\
$b_n$  & 27.5310&	0&	-0.8712&	0& 0 &	0 & 0 &	-0.9891 &	0 &	0 &	0 & 0 &	0 &	0 &	0 &0 &	0&	0&	0\\
$c_n$  & 0 &	0 &	0 &	0 & -2.6663 &	0 &	0.9998&	0 &	0 &	0 &	0 &	0 & 0 &	0 &	0 &	0 &	0 &	0&	0\\
\hline
\end{tabular}
\end{table}

\begin{figure*}[!ht]
\begin{tabular}{c c}
\centering
  \includegraphics[width=.6\textwidth]{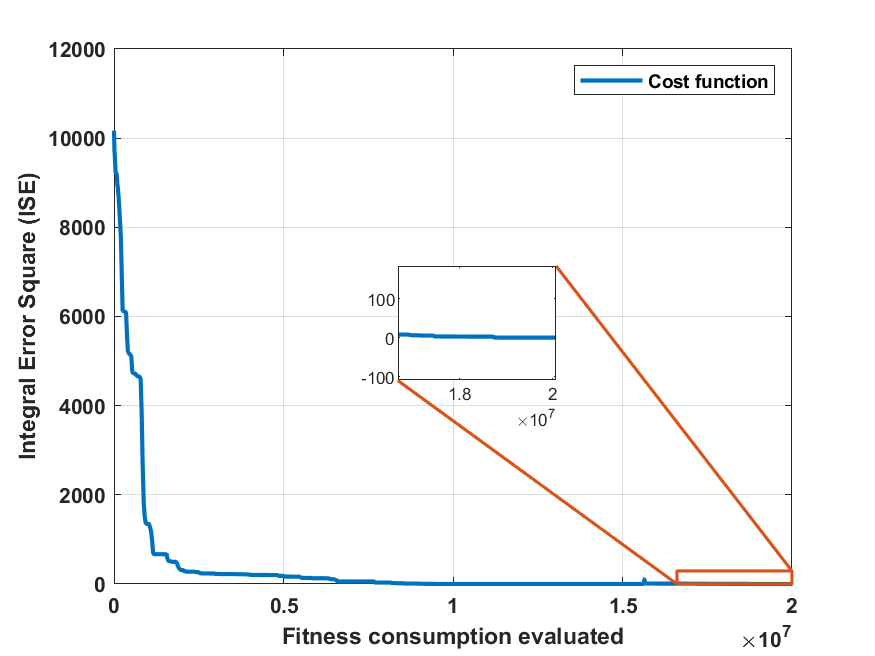}&
  \includegraphics[width=0.59\textwidth]{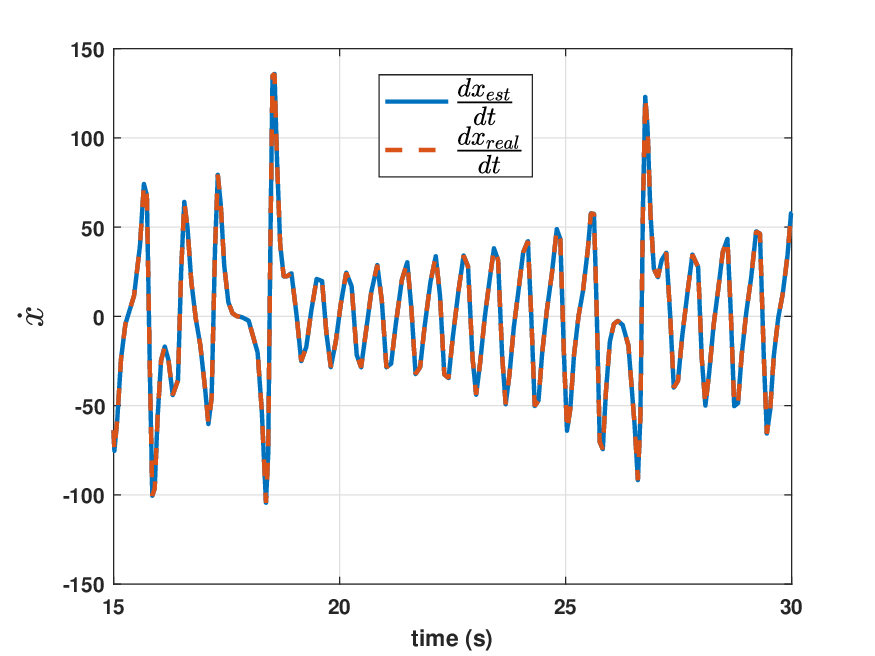}\\
    a)& b)\\
    \includegraphics[width=.6\textwidth]{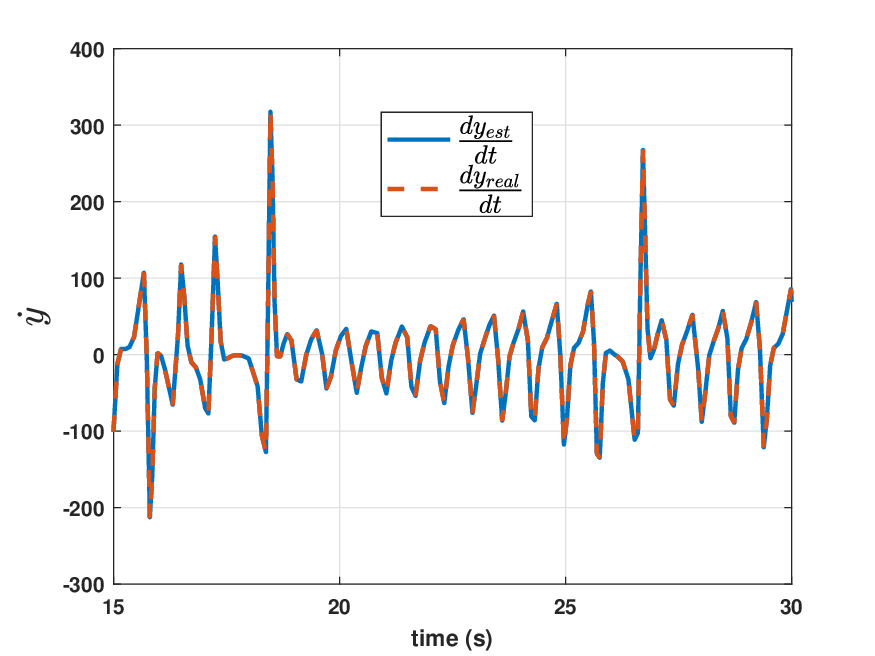}&
  \includegraphics[width=0.59\textwidth]{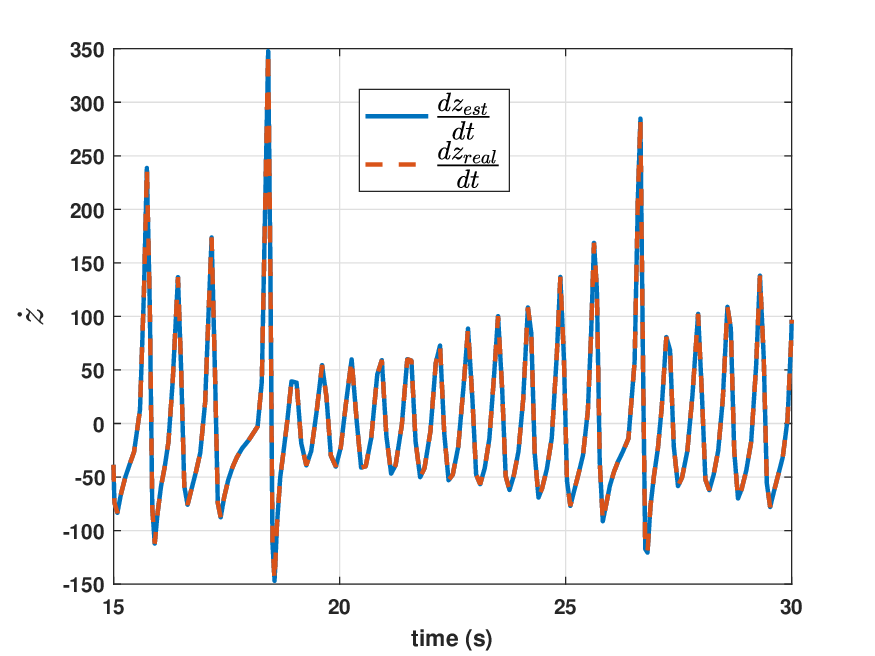}\\
  c)& d)
\end{tabular}
\caption{Results of the GA with dynamic limits for the Lorenz system. a) Performance of the algorithm. Comparison between the real and estimated values: b)  $\dot{x}_{est}$ vs $\dot{x}_{real}$ c) $\dot{y}_{est}$ vs $\dot{y}_{real}$, and  d) $\dot{z}_{est}$ vs $\dot{z}_{real}$.}
\centering
\label{LorFull}       
\end{figure*}

The performance of the algorithm is shown in Fig. \ref{LorFull} a), and a comparison between the real and estimated signals $\dot{x}$, $\dot{y}$ and $\dot{z}$ is presented in b), c), and d), respectively.
\clearpage

\begin{figure}[!ht]
\begin{tabular}{c c}
\centering
  \includegraphics[width=.6\textwidth]{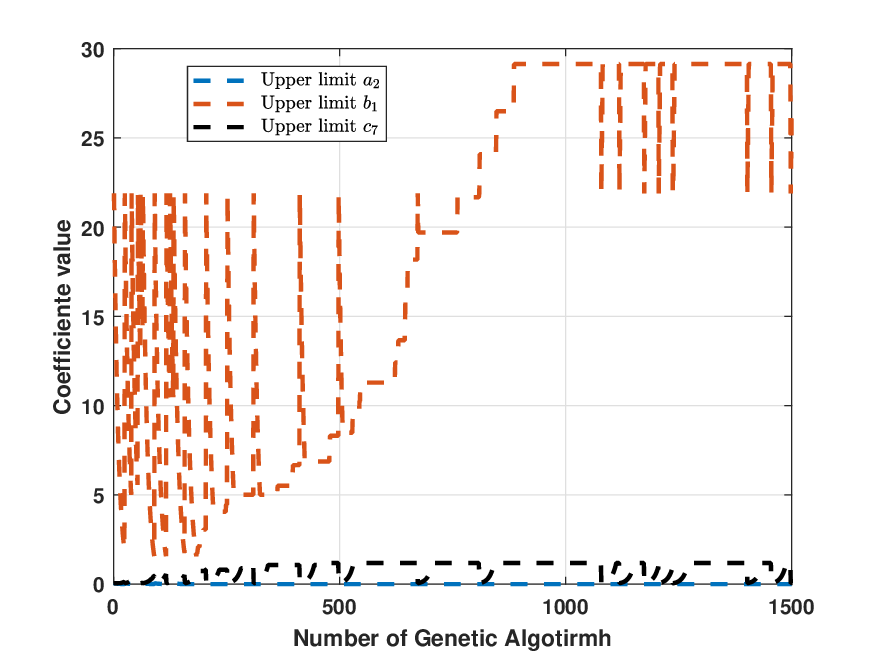}&
  \includegraphics[width=0.59\textwidth]{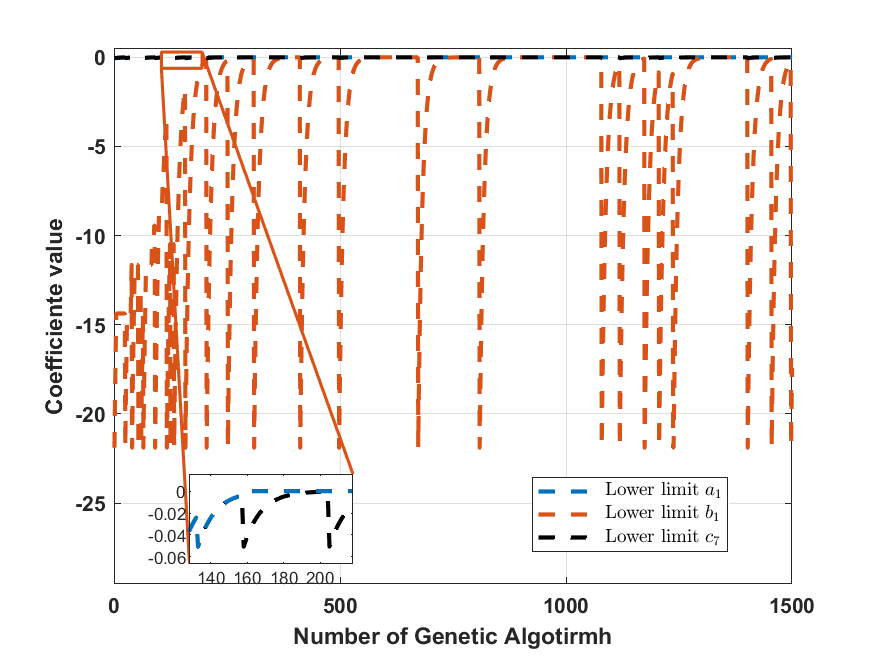}\\
    a)& b)\\
\end{tabular}
\caption{Evolution of dynamic search limits for dynamic system a) upper limits b) lower limits.}
\label{DLlor}
\end{figure}

 The estimation of $\dot{x}$ gives an ISE of 1.6 $\times 10^{-4}$, a MSE of 1.37 $\times 10^{-4}$, and a $R^2$ of 0.9999. For $\dot{y}$, the ISE was 0.3321, the MSE of 0.2833, and $R^2$ of 0.9999. Finally, for $\dot{z}$, an ISE of  3.82 $\times 10^{-4}$, MSE of 3.23 $\times 10^{-4}$ and $R^2$ of 0.9999. The evolution of some of the limits and the values of their respective coefficients are shown in Fig \ref{DLlor}. 

\begin{figure}[!ht]
\centering
\includegraphics[width=0.9\linewidth]{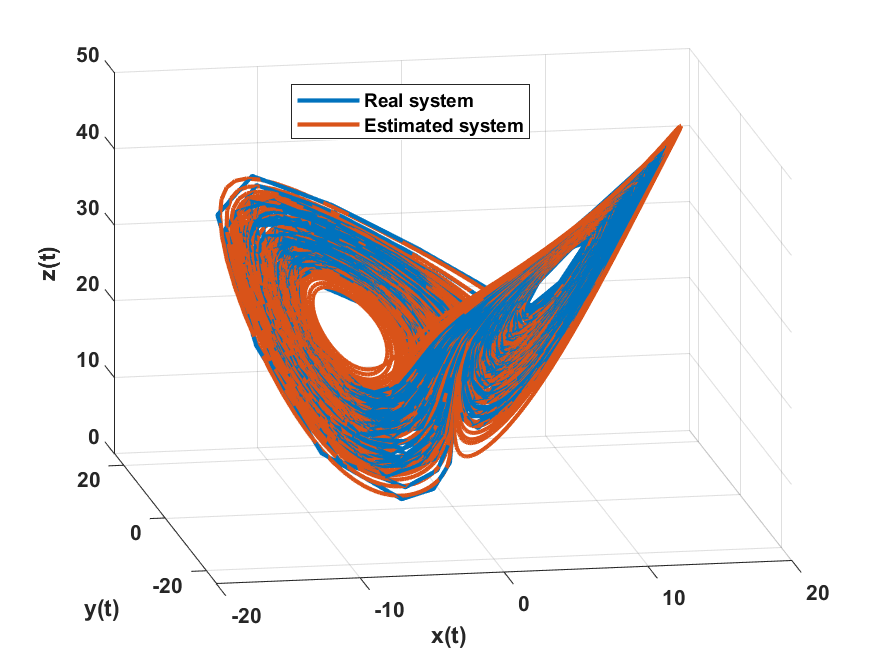}
\caption{Comparison between the real signal and estimated by the GA with the dynamic search limit for the Lorenz system.}
\label{LDLorInt}
\end{figure}

The real and estimated signals obtained by integration are shown in Fig \ref{LDLorInt}.

\clearpage

\section{Conclusions}

 The reconstruction of dynamical systems from data using genetic algorithms offers a powerful alternative for determining mathematical models in situations where the dynamic model is unknown or the laws governing the nonlinear dynamic system are unclear. This method provides an alternative dynamic model that behaves similarly to the original models. The main advantage of the proposed approach is that it allows the estimation of dynamic models with very little knowledge of the system. Even if the underlying system is unknown, only the time series of the original signals are required to determine the model. However, one of the main disadvantages of this method is that it lacks a fixed method for hyperparameter estimation and has higher computational costs than other methods. In addition, there is a possibility that no suitable model exists within the search space, in which case the algorithm will only provide the model with the lowest error.

Our results show that the genetic algorithm with fixed search limits was unable to solve the Lorenz attractor reconstruction problem; thus, we proposed a GA with dynamic search limits that we proved to be successful. Moreover, the modified algorithm outperformed the classical GA in the oscillator problem. This suggests that using dynamic limits in the search process can be a powerful tool for discovering dynamic models using metaheuristic algorithms.

\backmatter

\section*{Supplementary information}

Not applicable

\section*{Declarations}
This work was Funding by Consejo Nacional de Humanidades, Ciencias y Tecnologías
(CONAHCyT) of Mexico under Grant CF-2023-I-1496 and Dirección General de Asuntos del Personal
Académico (DGAPA)-National Autonomous University of Mexico (UNAM) under Project UNAM-
PAPIIT IA103325 and the postdoctoral fellowship DGAPA-UNAM for O.R.A.






\bibliography{sn-bibliography}

\end{document}